\documentclass{article}

\usepackage{arxiv}                       
\usepackage[utf8]{inputenc}
\usepackage[T1]{fontenc}
\usepackage{amsmath,amssymb,bm}
\usepackage{graphicx}
\usepackage{booktabs}
\usepackage{url}
\usepackage{microtype}
\usepackage[numbers,sort&compress]{natbib}
\usepackage[colorlinks=true,linkcolor=blue,citecolor=blue,urlcolor=blue]{hyperref}

\graphicspath{{./}{figures/}}

\title{Tropical Cyclone Forecasting via Latent Rectified Flow using
       Satellite Imagery and Atmospheric Fields}

\renewcommand{\shorttitle}{Latent Rectified Flow for Tropical Cyclone Forecasting}
\renewcommand{\headeright}{Preprint}
\renewcommand{\undertitle}{Preprint}

\hypersetup{
  pdftitle={Tropical Cyclone Forecasting via Latent Rectified Flow using
            Satellite Imagery and Atmospheric Fields},
  pdfauthor={Meheru Zannat, Sk. Md. Masudul Ahsan},
  pdfkeywords={tropical cyclone forecasting, rectified flow, flow matching,
               latent generative models, ERA5, GRIDSAT-B1, steering flow},
}

\author{%
  Meheru Zannat\\
  Department of Computer Science and Engineering\\
  Khulna University of Engineering \& Technology\\
  Khulna 9203, Bangladesh\\
  \texttt{meherujannat@gmail.com}\\
  \And
  Sk.~Md.~Masudul Ahsan\\
  Department of Computer Science and Engineering\\
  Khulna University of Engineering \& Technology\\
  Khulna 9203, Bangladesh\\
  \texttt{masudulahsan@gmail.com}\\
}

\date{\today}

\begin{document}
\maketitle

\begin{abstract}
Tropical cyclones are growing more destructive in a changing climate, and efficient forecasting of their structure and track has become a necessity. Deep generative models promise an alternative to computationally expensive numerical weather prediction (NWP), yet current systems produce either satellite imagery or atmospheric fields, never both; they need many sampling steps, putting them out of reach of modest hardware; and their storm tracks come from regression heads with no physical link to the generated atmosphere. This work presents a single-pass model that jointly forecasts GRIDSAT-B1 infrared imagery and four ERA5 atmospheric fields (U-wind, V-wind, air temperature, and surface pressure) out to nine hours. A five-channel variational autoencoder compresses each $5 \times 256 \times 256$ frame to a $4 \times 64 \times 64$ latent, and a conditional rectified-flow UNet with a factorized temporal-attention module predicts the next three frames from three past frames, their best-track coordinates, and timestamps. The model is then reward-fine-tuned (DRaFT) against a differentiable track error derived from the predicted winds through a steering-flow calculation. On held-out 2022 storms the model reaches 16.35\,dB PSNR and 0.759 SSIM, ahead of a reproduced cascaded-diffusion baseline at every lead time ($+0.84$\,dB at $+9$\,h) while sampling ${\sim}30\times$ faster (56\,ms vs.\ 1673\,ms). Track error at $+9$\,h is 62.4\,km, 15\% below the baseline, and a reward fine-tuning study demonstrates a further 8--11\% track-error reduction across sampler budgets.
\end{abstract}


\section{Introduction}
\label{sec:introduction}
Tropical cyclones strike the coasts of Bangladesh, eastern India, and Myanmar almost every year~\cite{nath2024forecasting}, and landfall on densely populated coastal areas causes massive casualties. Timely forecasts of cyclone structure and trajectory are essential for evacuation planning, but traditional numerical weather prediction (NWP) demands hours of supercomputer time per ensemble run, systematically under-predicts rapid intensification, and is too coarse ($9$--$25$\,km) to resolve inner-core structure~\cite{wang2024global}.

A complementary line of work learns cyclone evolution directly from historical observations with deep generative models. Denoising diffusion~\cite{ho2020denoising} and its latent variants~\cite{rombach2022high} now underpin global weather prediction (GenCast~\cite{price2023gencast}) as well as several cyclone-specific systems: cascaded diffusion over satellite imagery~\cite{nath2024forecasting}, video diffusion~\cite{ren2025improving}, and trajectory diffusion~\cite{zhang2025tcdiffuser}. Three limitations persist across these works. Each model produces either imagery or atmospheric fields, never both, so cross-channel physical consistency is never learned. Every forecast costs many sampling steps, which rules out rapid ensemble generation. And the track comes from a separate regression head, untied to the generated atmosphere.

This paper addresses the three limitations together with a \emph{multi-frame latent rectified flow}. A five-channel variational autoencoder (VAE) compresses each $5 \times 256 \times 256$ frame into a shared $4 \times 64 \times 64$ latent space. A rectified-flow~\cite{liu2022flow} UNet with factorized temporal attention learns the velocity field that generates the next three frames at 3-hour cadence in a single pass, conditioned on three past frames together with their best-track coordinates and timestamps.
The model is then fine-tuned with direct reward gradients (DRaFT~\cite{clark2024draft}) against a differentiable track error computed from the generated winds through a steering-flow calculation~\cite{chan2005global}. Against the cascaded-diffusion
baseline of Nath et al.~\cite{nath2024forecasting}, reproduced on the identical benchmark, the model comes out ahead at every lead time
while sampling ${\sim}30\times$ faster. Our contributions are:
\begin{enumerate}
    \item a single-pass generative model that forecasts satellite
          infrared imagery and ERA5 atmospheric fields \emph{jointly}
          for tropical cyclones;
    \item a multi-frame conditioning architecture that generates three
          future states at once and, with a single Euler step, matches
          the fifty-step diffusion baseline at ${\sim}62\times$ lower
          compute;
    \item a physics-derived steering-flow trajectory readout reaching
          17.5\,km at $+3$\,h and 62.4\,km at $+9$\,h with no track
          head; and
    \item a reward fine-tuning study showing direct reward gradients
          cut $+9$\,h track error by 8--11\% and transfer across
          sampler budgets.
\end{enumerate}

\section{Related Work}
\label{sec:related}

Operational cyclone forecasting still relies on NWP ensembles whose members each cost hours of supercomputer time.

Deterministic deep-learning emulators such as Pangu-Weather~\cite{bi2023pangu} and GraphCast~\cite{lam2023graphcast} match NWP at a fraction of the inference cost, but mean-regression training blurs convective structure and yields no native uncertainty.

Generative modelling of weather has expanded rapidly, with most systems built on denoising diffusion~\cite{ho2020denoising}. GenCast~\cite{price2023gencast} outperforms the leading operational ensemble, and for tropical cyclones the closest systems are the model of Nath et
al.~\cite{nath2024forecasting}, a three-stage cascade over infrared imagery; video diffusion~\cite{ren2025improving};
TC-Diffuser~\cite{zhang2025tcdiffuser}; TCP-Diffusion~\cite{huang2024tcp}; and satellite-to-atmosphere translation~\cite{ling2024estimating}. All commit to a single generated
modality and inherit diffusion's multi-step sampling.

Rectified flow~\cite{liu2022flow} learns a velocity field along the
\emph{straight-line} interpolant between noise and data, so that in the
limit of straight trajectories a single Euler step suffices. Latent
diffusion~\cite{rombach2022high} instead moves generation into the
lower-dimensional code of a separately trained VAE. To our knowledge, rectified flow has
not previously been applied to multi-channel tropical cyclone
forecasting. Table~\ref{tab:lit_gap} scores the closest systems along
the five design axes this paper combines.

\begin{table}[htbp]
    \centering
    \caption{Coverage of related work along the five design axes
             addressed jointly by this paper ($\checkmark$ full,
             $\sim$ partial, $\times$ absent).}
    \label{tab:lit_gap}
    \footnotesize
    \setlength{\tabcolsep}{3pt}
    \begin{tabular}{l c c c c c}
        \toprule
        \textbf{Method} & \textbf{5-ch} & \textbf{Latent} & \textbf{1-pass} & \textbf{Phys.} & \textbf{X-sens.} \\
        \midrule
        Nath et al.~\cite{nath2024forecasting}   & $\times$ & $\times$ & $\times$ & $\times$ & $\times$ \\
        Ren et al.~\cite{ren2025improving}       & $\times$ & $\times$ & $\times$ & $\times$ & $\times$ \\
        Ling et al.~\cite{ling2024estimating}    & $\sim$   & $\times$ & $\times$ & $\times$ & $\times$ \\
        TC-Diffuser~\cite{zhang2025tcdiffuser}   & $\sim$   & $\times$ & $\times$ & $\times$ & $\times$ \\
        TCP-Diffusion~\cite{huang2024tcp}        & $\times$ & $\times$ & $\times$ & $\times$ & $\times$ \\
        \textbf{Ours} & $\checkmark$ & $\checkmark$ & $\checkmark$ & $\checkmark$ & $\checkmark$ \\
        \bottomrule
    \end{tabular}
\end{table}

\section{Proposed Methodology}
\label{sec:method}

\begin{figure}[htbp]
    \centering
    \includegraphics[width=\textwidth]{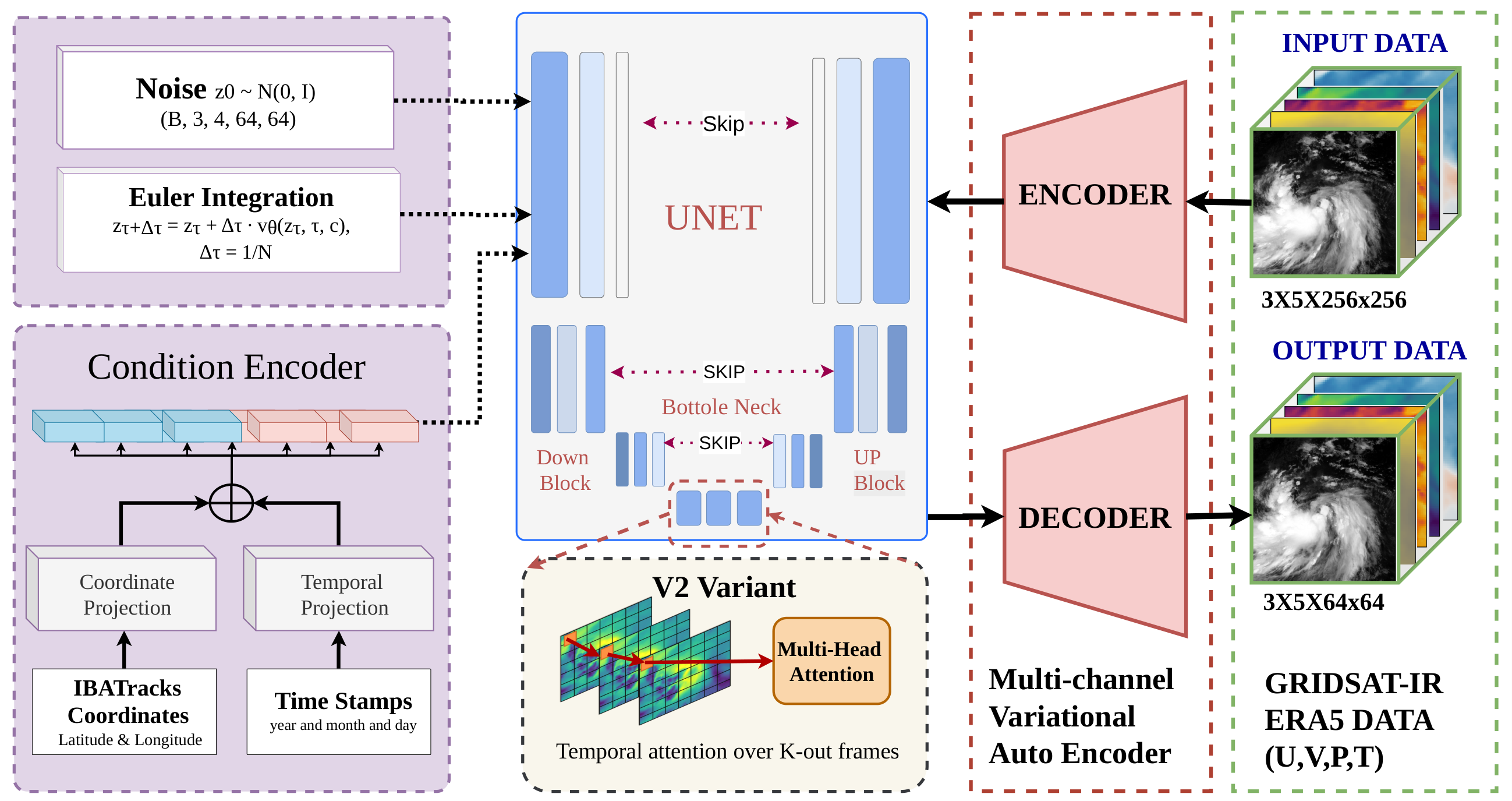}
    \caption{Two-stage framework: VAE compression (Stage~1),
             conditioned flow-matching UNet with temporal attention
             (Stage~2), and decoding with steering-flow trajectory
             readout.}
    \label{fig:framework}
\end{figure}

The proposed system (Fig.~\ref{fig:framework}) is a two-stage latent
generative model. Stage~1 is a multi-channel VAE that compresses each
five-channel frame to a compact latent frame; Stage~2 is a multi-frame
conditioned UNet with a factorized temporal attention module that learns the rectified-flow velocity field mapping a Gaussian prior to the
distribution of the next three latent frames. The storm trajectory is
calculated from the generated wind fields, and the model is finally
post-trained with direct reward gradients through the sampler.

\subsection{Problem Formulation}
\label{ssec:problem}

Let $\mathbf{x}_t \in \mathbb{R}^{5 \times 256 \times 256}$ denote a storm frame at time $t$, whose five channels are GRIDSAT-B1 infrared brightness temperature and ERA5 U-wind, V-wind, air
temperature (all at 850\,hPa), and surface pressure. Let
$\mathbf{c}_t = (\phi_t, \lambda_t)$ be the IBTrACS latitude/longitude at the same time and $\tau_t$ a calendar timestamp. Given $K_{\text{in}}$ past frames
$\{\mathbf{x}_{t-(K_{\text{in}}-1)}, \dots, \mathbf{x}_t\}$ with their coordinates and timestamps, the model must generate the next $K_{\text{out}}$ frames
$\{\mathbf{x}_{t+1}, \dots, \mathbf{x}_{t+K_{\text{out}}}\}$ at the native 3-hour cadence, i.e.\ sample from the conditional distribution
\begin{equation}
    p\big(\mathbf{x}_{t+1{:}t+K_{\text{out}}} \,\big|\,
          \mathbf{x}_{t-K_{\text{in}}+1{:}t},\,
          \mathbf{c},\, \tau\big).
    \label{eq:objective}
\end{equation}
The configuration throughout is $K_{\text{in}} = K_{\text{out}} = 3$: a 9-hour forecast from a 9-hour conditioning window.

\subsection{Data Preprocessing}
\label{ssec:preprocessing}

The following preprocessing steps are applied to the dataset:
\emph{(1)~Missing-value handling:} missing entries (NaN) in the 
GRIDSAT and ERA5 arrays are replaced with zeros.
\emph{(2)~Normalization:} all five channels are $z$-score normalized
with global per-channel statistics computed by Welford's online
algorithm over the training partition; the wide pressure standard
deviation ($\sigma \approx 521$\,Pa) and the $\pm 5$\,m/s wind scale
motivate the unbounded decoder of Section~\ref{ssec:vae}.

\emph{(3)~Dataset splitting:} the data are partitioned by calendar year, so each storm belongs to exactly one of the training, validation, and test partitions and the held-out years are identical across random seeds.

\emph{(4)~Augmentation:} at training time a horizontal flip with
probability $0.5$ is applied consistently to every frame and channel of
a window; a left-to-right flip preserves the poleward/equatorward
structure and the sense of Coriolis-driven rotation, which a rotation
would invert. Validation and test data are never augmented.

\subsection{Stage 1: Multi-Channel VAE}
\label{ssec:vae}

The VAE is a five-channel symmetric convolutional encoder-decoder
mapping $\mathbf{x} \in \mathbb{R}^{5 \times 256 \times 256}$ to a latent $\mathbf{z} \in \mathbb{R}^{4 \times 64 \times 64}$. The encoder $E_\phi$ downsamples from $256 \to 128 \to 64$ with strided $4 \times 4$ convolutions, each followed by GroupNorm and SiLU activation, growing the channel dimension from $5 \to 64 \to 128 \to 256$; two parallel $1 \times 1$ convolutions produce the posterior mean $\boldsymbol{\mu}$ and log-variance $\log\boldsymbol{\sigma}^2$, and the decoder $D_\theta$ mirrors the encoder. Training minimizes the $\beta$-weighted negative evidence lower bound (ELBO) with the KL weight annealed from $0$ to $10^{-4}$ over the first 20 epochs:
\begin{equation}
    \mathcal{L}_{\text{VAE}}
    = \mathbb{E}_{\mathbf{z} \sim q_\phi(\cdot|\mathbf{x})}
      \big[\|\mathbf{x} - D_\theta(\mathbf{z})\|_1\big]
    + \beta \, D_{\text{KL}}\!\big(q_\phi \| \mathcal{N}(\mathbf{0},\mathbf{I})\big).
    \label{eq:vae_loss}
\end{equation}
One deliberate departure from the standard latent-diffusion model is
that the decoder output is \emph{unbounded}: the customary $\tanh$ is
removed because $z$-scored reanalysis variables are naturally unbounded.
The $z$-scores of pressure regularly reach $\pm 3$, and clamping to
$[-1,1]$ destroyed more than half of the pressure dynamic range in
early experiments. A scale factor $s = 1/\sigma_{\mathbf{z}}$ estimated on the validation split rescales them to approximately unit variance for flow training.

\subsection{Conditioning Strategy}
\label{ssec:conditioning}

The generative stage sees three conditioning signals: the $K_{\text{in}}$
past-frame latents, the per-frame IBTrACS coordinates (latitude and longitude), and the calendar timestamps. A conditioning encoder embeds each past frame together with its timestamp into a per-frame context vector and aggregates the $K_{\text{in}}$ vectors; the coordinate pairs (normalized, flattened to $K_{\text{in}} \times 2$) pass through a small MLP whose output is added to the aggregated context. The combined context embedding is injected additively alongside the flow-time embedding at every UNet block.
Because the coordinates carry no spatial structure, they are kept \emph{out} of the VAE and enter only through this conditioning path.

\subsection{Rectified Flow UNet}
\label{ssec:flow}

Let $\mathbf{Z}_1$ denote the target latent block of shape
$K_{\text{out}} \cdot 4 \times 64 \times 64$ and
$\mathbf{Z}_0 \sim \mathcal{N}(\mathbf{0}, \mathbf{I})$ a Gaussian sample of the same shape. Rectified flow~\cite{liu2022flow} trains against the straight-line interpolant
\begin{equation}
    \mathbf{Z}_\tau = (1 - \tau)\,\mathbf{Z}_0 + \tau\,\mathbf{Z}_1,
    \qquad \tau \in [0, 1],
    \label{eq:interp}
\end{equation}
by regressing the network $v_\theta$ onto its constant velocity:
\begin{equation}
    \mathcal{L}_{\text{flow}}
    = \mathbb{E}_{\tau, \mathbf{Z}_0, (\mathbf{Z}_1, \mathbf{c})}
      \left[\big\|\,v_\theta(\mathbf{Z}_\tau, \tau, \mathbf{c})
              - (\mathbf{Z}_1 - \mathbf{Z}_0)\,\big\|_2^2\right].
    \label{eq:flow_loss}
\end{equation}
The velocity network is a latent-space UNet at $64 \times 64$: three
stride-2 downsamplings to an $8 \times 8$ bottleneck with channel
multipliers $(1,2,3,4) \times 128$, two ResNet blocks per stage,
eight-head self-attention at the $16 \times 16$ and $8 \times 8$
levels, and a mirrored skip-connected decoder (66.2\,M parameters).
Sampling is fixed-step Euler integration from $\tau = 0$ to $\tau = 1$:
\begin{equation}
    \mathbf{Z}_{\tau + \Delta\tau}
    = \mathbf{Z}_\tau
    + \Delta\tau \cdot v_\theta(\mathbf{Z}_\tau, \tau, \mathbf{c}),
    \quad \Delta\tau = 1/N_{\text{steps}}.
    \label{eq:euler}
\end{equation}
Because rectified training encourages near-straight sampling
trajectories, $N_{\text{steps}}$ can be very small; the measured sweep
(Section~\ref{ssec:inference}) peaks at $N_{\text{steps}} = 4$ with a
single step within $0.25$\,dB of the optimum.

\subsection{Factorized Temporal Attention Module}
\label{ssec:temporal}

Rather than channel-stacking the three output frames, the UNet is
shared across the $K_{\text{out}}$ output frames by folding the frame
axis into the batch axis, and a 1-D temporal self-attention module
mixes information across frames at the bottleneck. At the
$8 \times 8$ bottleneck, each spatial location forms one token per
output lead ($+3$, $+6$, $+9$\,h); the tokens are normalized, tagged with a learned per-lead slot embedding, mixed along the lead axis by multi-head self-attention, and added back residually. The module leaves the feature shape unchanged and adds almost no parameters; what it buys is that later output frames can attend to earlier ones.

\subsection{Steering-Flow Trajectory Readout}
\label{ssec:steering}

The trajectory is read out from the predicted U and V wind fields,
following the classical result that environmental steering explains
most of tropical-cyclone motion~\cite{chan2005global}. For each output
lead $k \in \{1,2,3\}$, the steering vector $(\bar{u}_k, \bar{v}_k)$ is the area mean of the
generated winds over the central $64 \times 64$ crop of the frame, denormalized to m/s.
The displacement over the 3-hour step $\Delta t = 10{,}800$\,s is
\begin{align}
    \Delta\phi_k    &= \frac{\bar{v}_k \, \Delta t}{111{,}000},
    \label{eq:steering_lat}\\[2pt]
    \Delta\lambda_k &= \frac{\bar{u}_k \, \Delta t}
                            {111{,}000 \cdot \cos(\phi_t + \Delta\phi_k/2)},
    \label{eq:steering_lon}
\end{align}
where $\phi_t$ is the current latitude. Track error is measured as the great-circle separation of the forecast position from the observed one; because the position derives entirely from the generated atmosphere, it doubles as a probe of the physical consistency of the generated fields.

\subsection{Reward Fine-Tuning through the Sampler}
\label{ssec:draft}

Velocity regression~\eqref{eq:flow_loss} is a surrogate objective: every evaluation metric is computed on the multi-step \emph{sampled} forecast, which the training loss never sees. We close this gap with
direct reward fine-tuning (DRaFT~\cite{clark2024draft}). The
deterministic Euler sampler is unrolled, and only the final $K = 2$ of
$N = 10$ steps and the frozen-VAE decode carry gradient (DRaFT-$K$
truncation). The reward is a differentiable version of the target
metric, a smooth steering-track error built
from~\eqref{eq:steering_lat} and~\eqref{eq:steering_lon}, combined with a
pixel-MSE guard and an $L_2$-SP anchor to the pre-fine-tuning weights
to prevent reward hacking. Progress is monitored on a seeded 64-window validation ruler; all reported claims use the held-out test set.

\section{Experiments and Results}
\label{sec:experiments}

\subsection{Dataset}
\label{ssec:dataset}

All experiments use the Satellite ERA5 Tropical Cyclone Dataset
(SETCD), assembled from three open archives aligned by storm identifier and timestamp: GRIDSAT-B1~\cite{knapp2011gridsat} infrared brightness temperature, four ERA5 fields~\cite{hersbach2020era5} (850\,hPa U-wind, V-wind, air temperature, and surface pressure) at $0.25^{\circ}$ resolution and 3-hour cadence, and
IBTrACS~v04r01~\cite{knapp2010ibtracs} coordinates. Every
sample is a $256 \times 256$ storm-centred window.
Table~\ref{tab:basins} lists the basins covered. The calendar-year
split is training 2012--2020 (24{,}625 windows), validation 2021
(2{,}852), and test 2022 (1{,}732). For the cross-sensor experiment
(Section~\ref{ssec:cross_dataset}) we assembled \emph{SETCD-WP}: 500 multi-frame windows spanning nine Western-Pacific storms (eight from 2019, Himawari-8; one from 1978--79, GMS-1), with infrared imagery from the Digital Typhoon archive and matching ERA5 fields.

\begin{table}[htbp]
    \centering
    \caption{Basins in SETCD with the primary geostationary sensor
             feeding GRIDSAT-B1 and a representative landfall
             sub-region (distinct IBTrACS storms, 2012--2022; N/S =
             hemisphere).}
    \label{tab:basins}
    \footnotesize
    \setlength{\tabcolsep}{2pt}
    \begin{tabular}{c l l l r}
        \toprule
        & \textbf{Basin} & \textbf{Primary sensor} & \textbf{Sub-region} & \textbf{Storms} \\
        \midrule
        N & Western Pacific & Himawari-8/9      & East/SE Asia         & 336 \\
        N & North Atlantic  & GOES-East         & US East/Caribbean    & 247 \\
        N & Eastern Pacific & GOES-West         & Central America      & 189 \\
        N & North Indian    & INSAT-3D/Met-IODC & S Asia (Bay of Bengal) & 100 \\
        \midrule
        S & South Indian    & Meteosat-9        & Madagascar/Mascarene & 181 \\
        S & South Pacific   & Himawari-9/GOES-W & Australia/Oceania    & 107 \\
        \midrule
        \multicolumn{4}{l}{\textbf{Total}} & \textbf{1160} \\
        \bottomrule
    \end{tabular}
\end{table}
\subsection{Evaluation Metrics}
\label{ssec:metrics}
Forecast quality is assessed along three axes at each lead time. \emph{Pixel fidelity}: mean absolute error (MAE), peak signal-to-noise ratio (PSNR), and structural similarity (SSIM), each computed over the five output channels. \emph{Probabilistic calibration}: the continuous ranked probability score (CRPS), the standard ensemble-verification measure~\cite{price2023gencast}, over ten stochastic samples per window drawn by re-sampling the initial noise. \emph{Trajectory accuracy}: the distance in km between the steering-flow position and the IBTrACS position along the great circle joining them.

\subsection{Baseline Methods}
\label{ssec:baselines}

The external baseline is the cascaded-diffusion model of Nath et
al.~\cite{nath2024forecasting}, which we re-implemented and trained on the same SETCD split with the identical evaluation pipeline ($64 \to 128 \to 256$ cascade trained for 60/50/40 epochs, classifier-free guidance 3.0, 50-step denoising diffusion implicit model (DDIM) sampling).
The ablation study (Section~\ref{ssec:ablations}) further compares
two internal variants of the proposed backbone that replace the
temporal-attention module, V1 (channel-stacked outputs, additive
fusion, no temporal mixing) and V3 (cross-attention to condition
tokens), plus a single-frame model rolled out autoregressively.

\subsection{Implementation Details}
\label{ssec:implementation}

All models are implemented in PyTorch with BF16 mixed precision, exponential-moving-average (EMA) weight tracking (decay $0.9999$), and gradient clipping at $1.0$. Adam is used with weight decay $10^{-5}$ and a learning rate of $10^{-4}$, decayed by a cosine schedule to $5 \times 10^{-5}$ after five warm-up epochs. The VAE trains for 60 epochs; the UNet trains for 100 epochs with a per-step batch of 4 accumulated over four steps, an effective batch of 16. Every experiment fits on a single GPU (NVIDIA RTX A6000 48\,GB; each configuration also reproduces on a 16\,GB consumer card), and all experiments combined consumed roughly 400 GPU-hours. Unless noted, evaluation uses 50-step Euler sampling.

\subsection{Quantitative Results}
\label{ssec:quantitative}

Table~\ref{tab:main_results} reports per-leadtime test-set metrics
against the reproduced cascaded-diffusion baseline. The proposed model achieves higher PSNR at every lead: $+0.15$\,dB at $+3$\,h,
widening monotonically to $+0.84$\,dB at $+9$\,h. The steering-flow
track error is 17.5\,km at $+3$\,h and 62.4\,km at $+9$\,h, 15\% below the baseline at the longest lead. CRPS, in contrast, remains $1.5$--$1.8\times$ the cascaded baseline's, a structural property of near-deterministic flow sampling discussed below. A VAE-ceiling diagnostic bounds any latent model on this data at 33.04\,dB PSNR / 0.982 SSIM (surface pressure lowest, 29.20\,dB), confirming that the generative stage, not the autoencoder, is the bottleneck of the system.

\begin{table}[htbp]
    \centering
    \caption{Per-leadtime test-set metrics versus the reproduced
             cascaded-diffusion baseline (50-step sampling). MAE is in
             $z$-scored units. Best per metric and lead in bold.}
    \label{tab:main_results}
    \small
    \setlength{\tabcolsep}{3pt}
    \begin{tabular}{l l c c c c c}
        \toprule
        \textbf{Model} & \textbf{Lead} & \textbf{PSNR} & \textbf{SSIM} & \textbf{MAE} & \textbf{CRPS} & \textbf{Track (km)} \\
        \midrule
        \textbf{Ours} & $+3$\,h & \textbf{17.36} & \textbf{0.770} & 0.198 & 0.130 & \textbf{17.5} \\
                      & $+6$\,h & \textbf{16.39} & \textbf{0.753} & \textbf{0.221} & 0.148 & \textbf{41.0} \\
                      & $+9$\,h & \textbf{15.48} & \textbf{0.733} & \textbf{0.243} & 0.166 & \textbf{62.4} \\
        \midrule
        Nath et al.~\cite{nath2024forecasting}
                      & $+3$\,h & 17.21 & 0.768 & 0.198 & \textbf{0.085} & 17.6 \\
                      & $+6$\,h & 15.86 & 0.736 & 0.231 & \textbf{0.089} & 43.0 \\
                      & $+9$\,h & 14.64 & 0.706 & 0.266 & \textbf{0.093} & 73.4 \\
        \bottomrule
    \end{tabular}
\end{table}

\emph{Reward fine-tuning.} Track-focused DRaFT fine-tuning
(Section~\ref{ssec:draft}) reduces the $+9$\,h track error by
$11.0\%$ at the 50-step sampler budget (held out from fine-tuning, so
the gain is not budget adaptation) and by $8.3\%$ at the 10-step
training budget, with SSIM improved at every lead and PSNR essentially
unchanged. Two negative results
complete the study. Policy-gradient fine-tuning with group-relative
policy optimization (GRPO~\cite{liu2025flowgrpo}) left every metric flat within noise; the
group-relative advantage signal is vanishingly small in a
${\sim}$49k-dimensional latent at academic batch sizes. A correctly
specified all-channel CRPS reward also stayed flat, which indicates the
ensemble under-dispersion of a near-deterministic flow sampler is
structural rather than trainable; closing it requires stochastic
inference sampling or CRPS-trained distillation~\cite{jacq2026rmmd}.

\subsection{Qualitative Results}
\label{ssec:qualitative}

\begin{figure}[htbp]
    \centering
    \includegraphics[width=\textwidth]{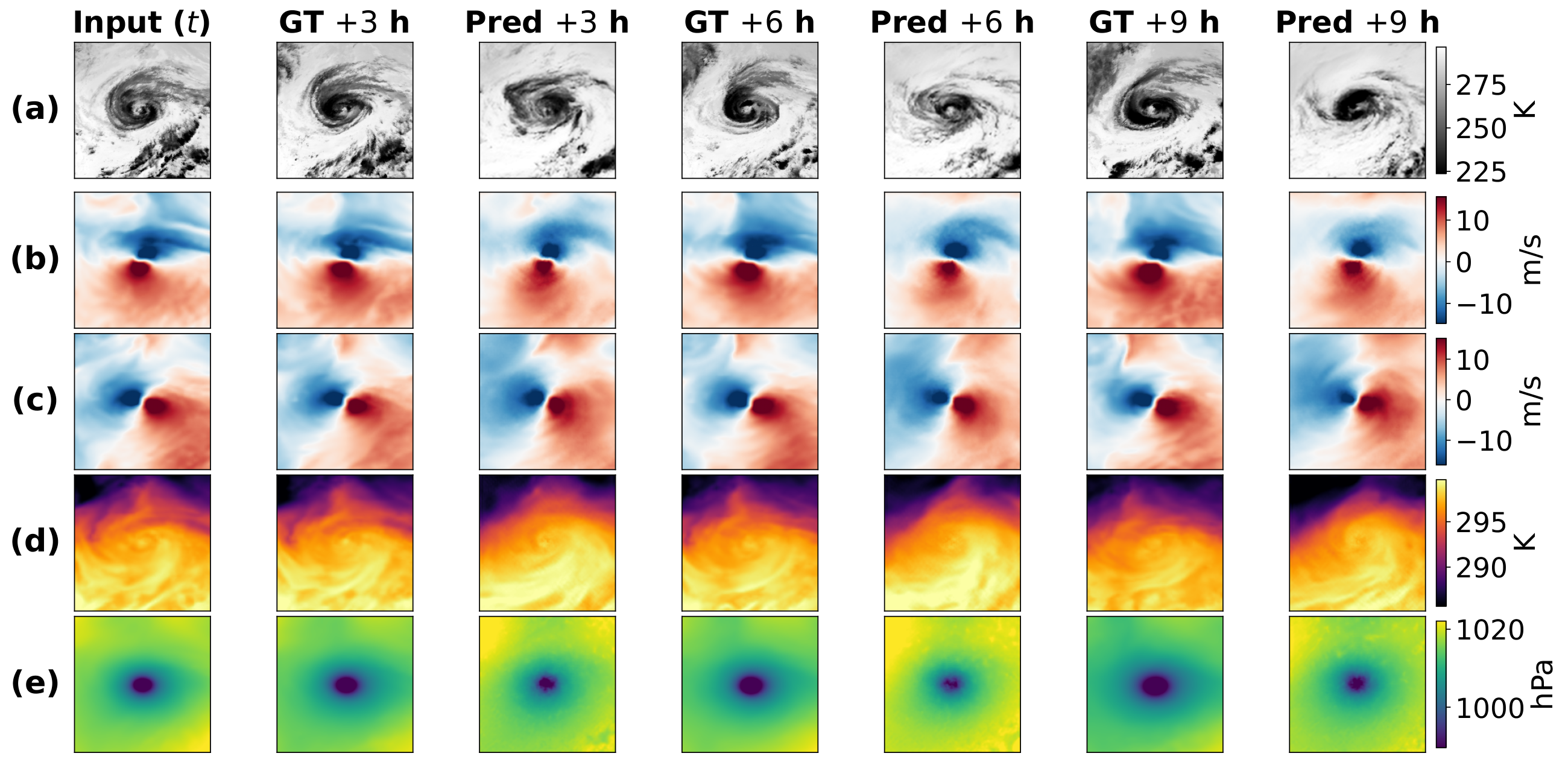}
    \caption{All-channel forecast for a held-out 2022 test storm: the
             last input frame, then ground truth (GT) and prediction at
             the $+3$, $+6$, and $+9$\,h leads.
             (a)~GRIDSAT-B1 infrared brightness temperature.
             (b)~850\,hPa U-wind. (c)~850\,hPa V-wind.
             (d)~850\,hPa air temperature. (e)~Surface pressure.}
    \label{fig:forecast_allch}
\end{figure}

Fig.~\ref{fig:forecast_allch} shows an all-channel forecast for a
held-out test storm. The model captures the central dense overcast and
spiral rain-band structure in the infrared, and the generated wind,
temperature, and pressure fields remain spatially coherent with the
predicted cloud structure at every lead. The recurring failure modes are cloud-top texture smoothing at $+9$\,h, 2--5\,hPa under-prediction of
the deepest central pressures, and $2\times$ the average track error on
rapidly recurving storms.

\subsection{Inference Analysis}
\label{ssec:inference}

Table~\ref{tab:sampler_budget} reports the measured sampler-budget
sweep (mean $\pm$ s.d.\ across three training seeds). Quality
\emph{peaks} at four Euler steps (16.50\,dB) and decreases slightly
beyond: extra integration adds cost without benefit. A single Euler step retains 16.25\,dB, only 0.25\,dB below the optimum, at 27\,ms per forecast. The baseline needs fifty DDIM steps and 1673\,ms to reach 16.19\,dB, so the proposed model at four steps is both $+0.31$\,dB better and ${\sim}30\times$ faster, and at one step it matches the baseline's best quality at ${\sim}62\times$ lower compute. That margin is what makes large real-time forecast ensembles affordable on commodity hardware.

\begin{table}[htbp]
    \centering
    \caption{Sampler-budget sweep on the test set (ours: mean $\pm$
             s.d.\ across three seeds).}
    \label{tab:sampler_budget}
    \small
    \setlength{\tabcolsep}{5pt}
    \begin{tabular}{l c c c r}
        \toprule
        \textbf{Model} & \textbf{Steps} & \textbf{PSNR (dB)} & \textbf{SSIM} & \textbf{Time (ms)} \\
        \midrule
        Ours      & 1  & $16.25 \pm 0.08$ & 0.724 & \textbf{27.1} \\
                  & 4  & $\mathbf{16.50 \pm 0.09}$ & \textbf{0.740} & 56.3 \\
                  & 16 & $16.25 \pm 0.14$ & 0.739 & 119.0 \\
                  & 50 & $16.12 \pm 0.13$ & 0.732 & 330.5 \\
        \midrule
        Nath et al.~\cite{nath2024forecasting}
                  & 25 & 15.76 & 0.723 & 848.3 \\
                  & 50 & 16.19 & 0.721 & 1673.3 \\
        \bottomrule
    \end{tabular}
\end{table}

\subsection{Cross-Dataset Validation}
\label{ssec:cross_dataset}

The unchanged checkpoint was evaluated zero-shot on SETCD-WP
(Table~\ref{tab:cross_dataset}); because the Western Pacific is well
represented in training, this isolates the sensor and processing shift
from the storm climatology.
Degradation is graceful: every metric worsens smoothly with lead, as it does in-domain, and SSIM stays substantial (0.68--0.60), so the model transfers cyclone dynamics rather than collapsing. A per-storm breakdown traces most of the absolute gap to the two (of nine) storms whose windows cross the high terrain of Indochina, where ERA5 surface pressure falls to 857\,hPa, roughly $-29$ standard deviations relative to the ocean-dominated training statistics, and the pressure channel becomes unreconstructable. The seven over-ocean storms
reach 11.1/9.8/8.9\,dB, only ${\sim}6.5$\,dB below in-domain; a terrain-insensitive mean-sea-level pressure channel would eliminate the blow-up without retraining.

\begin{table}[htbp]
    \centering
    \caption{Zero-shot cross-sensor generalization: in-domain
             GRIDSAT-merged imagery versus the single-satellite
             Digital Typhoon SETCD-WP set.}
    \label{tab:cross_dataset}
    \small
    \setlength{\tabcolsep}{5pt}
    \begin{tabular}{l l c c c}
        \toprule
        \textbf{Test set} & \textbf{Lead} & \textbf{PSNR (dB)} & \textbf{SSIM} & \textbf{Track (km)} \\
        \midrule
        SETCD       & $+3$\,h & 17.36 & 0.770 & 17.5 \\
        (in-domain) & $+6$\,h & 16.39 & 0.753 & 41.0 \\
                    & $+9$\,h & 15.48 & 0.733 & 62.4 \\
        \midrule
        SETCD-WP    & $+3$\,h & 10.00 & 0.677 & 57.1 \\
        (zero-shot) & $+6$\,h & 8.76  & 0.634 & 132.5 \\
                    & $+9$\,h & 7.93  & 0.597 & 218.1 \\
        \bottomrule
    \end{tabular}
\end{table}

\subsection{Ablation Study}
\label{ssec:ablations}

Table~\ref{tab:ablation_summary} collects every reported training run.
\emph{Temporal attention wins:} at fixed context and
objective, the proposed factorized temporal-attention backbone beats
the V1 (channel-stack) and V3 (cross-attention) alternatives by
0.10--0.23\,dB overall, consistently at every lead. \emph{Multi-frame conditioning is necessary:} a single-frame
model rolled out autoregressively, even with ground-truth coordinates
supplied at every step, matches direct generation at $+3$\,h but loses
1.09\,dB at $+6$\,h and 1.38\,dB at $+9$\,h; $+5.65$\,dB of the $+9$\,h gap sits in the temperature channel alone, because rollout destroys the slow dynamical evolution that direct multi-frame conditioning preserves.
\emph{Physics-informed auxiliary losses do not
help:} adding mass-continuity, geostrophic-balance, and
track-consistency penalties on the decoded prediction changes PSNR by $-0.23$\,dB at design weights and degrades it monotonically at $10\times$ weights; the constraints appear to be
implicit in the joint five-channel objective already.
\emph{Reflow (2-RF) actively hurts:} the second-pass rectification
lands 0.42\,dB below the one-pass model at $2.3\times$ the training
cost, so we use the 1-RF formulation throughout. (The single-frame model's higher overall PSNR reflects a strictly easier $+3$\,h-only problem.)

\begin{table}[htbp]
    \centering
    \caption{Ablation summary (overall PSNR/SSIM aggregated across
             leads).}
    \label{tab:ablation_summary}
    \small
    \setlength{\tabcolsep}{2.5pt}
    \begin{tabular}{l c c c c}
        \toprule
        \textbf{Configuration} & $K$ & \textbf{Physics} & \textbf{PSNR (dB)} & \textbf{SSIM} \\
        \midrule
        Single-frame (V1)           & 1 & off        & 18.71 & 0.810 \\
        Single-frame + physics      & 1 & $1\times$  & 18.67 & 0.808 \\
        \midrule
        Multi-frame, V1             & 3 & off        & 16.25 & 0.753 \\
        Multi-frame, V1 + physics   & 3 & $1\times$  & 16.02 & 0.758 \\
        Multi-frame, V1 + physics   & 3 & $10\times$ & 15.71 & 0.755 \\
        Multi-frame, V1, reflow     & 3 & off        & 15.83 & 0.751 \\
        \textbf{Proposed (temporal attn.)} & 3 & off & $\mathbf{16.35}$ & $\mathbf{0.759}$ \\
        Multi-frame, V3             & 3 & off        & 16.12 & 0.752 \\
        \bottomrule
    \end{tabular}
\end{table}

\section{Conclusion}
\label{sec:conclusion}

We presented a multi-frame latent rectified-flow model that jointly
forecasts satellite infrared imagery and four ERA5 atmospheric fields
for tropical cyclones at a nine-hour horizon, in a single forward pass
on one commodity GPU. On held-out 2022 storms spanning all major
basins, the model beats a reproduced cascaded-diffusion baseline at
every lead time ($+0.84$\,dB at $+9$\,h) while sampling
${\sim}30\times$ faster. The storm track error read out physically from the
generated winds reaches 62.4\,km at $+9$\,h, 15\% below the baseline, and direct reward fine-tuning through the sampler trims track error by a further 8--11\% without degrading fidelity. Ablations show the temporal-attention backbone and multi-frame conditioning are genuinely responsible for the long-lead gains. The main limitations are pixel-MSE texture smoothing, structurally under-dispersed ensembles, and the terrain-sensitive surface-pressure channel; these point to perceptual objectives, stochastic inference sampling, and a mean-sea-level pressure channel as future work, alongside extending the horizon to 24--48\,h.


\bibliographystyle{unsrtnat}
\bibliography{references}

\begin{thebibliography}{19}
\providecommand{\natexlab}[1]{#1}
\providecommand{\url}[1]{\texttt{#1}}
\expandafter\ifx\csname urlstyle\endcsname\relax
  \providecommand{\doi}[1]{doi: #1}\else
  \providecommand{\doi}{doi: \begingroup \urlstyle{rm}\Url}\fi

\bibitem[Nath et~al.(2024)Nath, Shukla, Wang, and
  Quilodr{\'a}n-Casas]{nath2024forecasting}
Pritthijit Nath, Pancham Shukla, Shuai Wang, and C{\'e}sar Quilodr{\'a}n-Casas.
\newblock Forecasting tropical cyclones with cascaded diffusion models.
\newblock In \emph{ICLR Workshop on Tackling Climate Change with Machine
  Learning}, Vienna, Austria, 2024.
\newblock \doi{10.48550/arXiv.2310.01690}.

\bibitem[Wang et~al.(2024)Wang, Chen, Liu, Han, and Li]{wang2024global}
Xinyu Wang, Kang Chen, Lei Liu, Tao Han, and Bin Li.
\newblock Global tropical cyclone intensity forecasting with multi-modal
  multi-scale causal autoregressive model.
\newblock In \emph{Advances in Neural Information Processing Systems},
  volume~37, Vancouver, Canada, 2024.

\bibitem[Ho et~al.(2020)Ho, Jain, and Abbeel]{ho2020denoising}
Jonathan Ho, Ajay Jain, and Pieter Abbeel.
\newblock Denoising diffusion probabilistic models.
\newblock In \emph{Advances in Neural Information Processing Systems},
  volume~33, pages 6840--6851, 2020.

\bibitem[Rombach et~al.(2022)Rombach, Blattmann, Lorenz, Esser, and
  Ommer]{rombach2022high}
Robin Rombach, Andreas Blattmann, Dominik Lorenz, Patrick Esser, and Bj{\"o}rn
  Ommer.
\newblock High-resolution image synthesis with latent diffusion models.
\newblock In \emph{Proceedings of the IEEE/CVF Conference on Computer Vision
  and Pattern Recognition}, pages 10684--10695, New Orleans, LA, USA, 2022.
\newblock \doi{10.1109/CVPR52688.2022.01042}.

\bibitem[Price et~al.(2025)Price, Sanchez-Gonzalez, Alet, Andersson, El-Kadi,
  Masters, Ewalds, Stott, Mohamed, Battaglia, Lam, and
  Willson]{price2023gencast}
Ilan Price, Alvaro Sanchez-Gonzalez, Ferran Alet, Tom~R. Andersson, Andrew
  El-Kadi, Dominic Masters, Timo Ewalds, Jacklynn Stott, Shakir Mohamed, Peter
  Battaglia, Remi Lam, and Matthew Willson.
\newblock Probabilistic weather forecasting with machine learning.
\newblock \emph{Nature}, 637:\penalty0 84--90, 2025.
\newblock \doi{10.1038/s41586-024-08252-9}.

\bibitem[Ren et~al.(2025)Ren, Nath, Shukla, and
  Quilodr{\'a}n-Casas]{ren2025improving}
Zhibo Ren, Pritthijit Nath, Pancham Shukla, and C{\'e}sar Quilodr{\'a}n-Casas.
\newblock Improving tropical cyclone forecasting with video diffusion models.
\newblock In \emph{ICLR Workshop on Tackling Climate Change with Machine
  Learning}, Singapore, 2025.
\newblock \doi{10.48550/arXiv.2501.16003}.

\bibitem[Zhang et~al.(2025)Zhang, Mu, Huang, Zhang, and
  Bai]{zhang2025tcdiffuser}
Sijia Zhang, Pan Mu, Cheng Huang, Jinglin Zhang, and Cong Bai.
\newblock {TC-Diffuser}: Bi-condition multi-modal diffusion for tropical
  cyclone forecasting.
\newblock In \emph{Proceedings of the AAAI Conference on Artificial
  Intelligence}, volume~39, pages 32099--32107, Philadelphia, PA, USA, 2025.
\newblock \doi{10.1609/aaai.v39i1.32099}.

\bibitem[Liu et~al.(2023)Liu, Gong, and Liu]{liu2022flow}
Xingchao Liu, Chengyue Gong, and Qiang Liu.
\newblock Flow straight and fast: Learning to generate and transfer data with
  rectified flow.
\newblock In \emph{International Conference on Learning Representations},
  Kigali, Rwanda, 2023.

\bibitem[Clark et~al.(2024)Clark, Vicol, Swersky, and Fleet]{clark2024draft}
Kevin Clark, Paul Vicol, Kevin Swersky, and David~J. Fleet.
\newblock Directly fine-tuning diffusion models on differentiable rewards.
\newblock In \emph{International Conference on Learning Representations},
  Vienna, Austria, 2024.
\newblock \doi{10.48550/arXiv.2309.17400}.

\bibitem[Chan(2005)]{chan2005global}
Johnny C.~L. Chan.
\newblock The physics of tropical cyclone motion.
\newblock \emph{Annual Review of Fluid Mechanics}, 37:\penalty0 99--128, 2005.
\newblock \doi{10.1146/annurev.fluid.37.061903.175702}.

\bibitem[Bi et~al.(2023)Bi, Xie, Zhang, Chen, Gu, and Tian]{bi2023pangu}
Kaifeng Bi, Lingxi Xie, Hengheng Zhang, Xin Chen, Xiaotao Gu, and Qi~Tian.
\newblock Accurate medium-range global weather forecasting with {3D} neural
  networks.
\newblock \emph{Nature}, 619:\penalty0 533--538, 2023.
\newblock \doi{10.1038/s41586-023-06185-3}.

\bibitem[Lam et~al.(2023)Lam, Sanchez-Gonzalez, Willson, Wirnsberger,
  Fortunato, Alet, et~al.]{lam2023graphcast}
Remi Lam, Alvaro Sanchez-Gonzalez, Matthew Willson, Peter Wirnsberger, Meire
  Fortunato, Ferran Alet, et~al.
\newblock Learning skillful medium-range global weather forecasting.
\newblock \emph{Science}, 382\penalty0 (6677):\penalty0 1416--1421, 2023.
\newblock \doi{10.1126/science.adi2336}.

\bibitem[Huang et~al.(2024)Huang, Mu, Bai, and Watson]{huang2024tcp}
Cheng Huang, Pan Mu, Cong Bai, and Peter A.~G. Watson.
\newblock {TCP-Diffusion}: A multi-modal diffusion model for global tropical
  cyclone precipitation forecasting with change awareness.
\newblock In \emph{Proceedings of the International Conference on Machine
  Learning}, Vienna, Austria, 2024.
\newblock \doi{10.48550/arXiv.2410.13175}.

\bibitem[Ling et~al.(2024)Ling, Nath, and
  Quilodr{\'a}n-Casas]{ling2024estimating}
Zhangyue Ling, Pritthijit Nath, and C{\'e}sar Quilodr{\'a}n-Casas.
\newblock Estimating atmospheric variables from digital typhoon satellite
  images via conditional denoising diffusion models.
\newblock \emph{arXiv preprint arXiv:2409.07961}, 2024.
\newblock \doi{10.48550/arXiv.2409.07961}.

\bibitem[Knapp et~al.(2011)Knapp, Ansari, Bain, Bourassa, Dickinson, Funk,
  Helms, Hennon, Holmes, Huffman, Kossin, Lee, Loew, and
  Magnusdottir]{knapp2011gridsat}
Kenneth~R. Knapp, Steve Ansari, Caroline~L. Bain, Mark~A. Bourassa, Michael~J.
  Dickinson, Chris Funk, Chip~N. Helms, Christopher~C. Hennon, Christopher~D.
  Holmes, George~J. Huffman, James~P. Kossin, Hai-Tien Lee, Alexander Loew, and
  Gudrun Magnusdottir.
\newblock Globally gridded satellite observations for climate studies.
\newblock \emph{Bulletin of the American Meteorological Society}, 92\penalty0
  (7):\penalty0 893--907, 2011.
\newblock \doi{10.1175/2011BAMS3039.1}.

\bibitem[Hersbach et~al.(2020)]{hersbach2020era5}
Hans Hersbach et~al.
\newblock The {ERA5} global reanalysis.
\newblock \emph{Quarterly Journal of the Royal Meteorological Society},
  146\penalty0 (730):\penalty0 1999--2049, 2020.
\newblock \doi{10.1002/qj.3803}.

\bibitem[Knapp et~al.(2010)Knapp, Kruk, Levinson, Diamond, and
  Neumann]{knapp2010ibtracs}
Kenneth~R. Knapp, Michael~C. Kruk, David~H. Levinson, Howard~J. Diamond, and
  Charles~J. Neumann.
\newblock The international best track archive for climate stewardship
  ({IBTrACS}): Unifying tropical cyclone data.
\newblock \emph{Bulletin of the American Meteorological Society}, 91\penalty0
  (3):\penalty0 363--376, 2010.
\newblock \doi{10.1175/2009BAMS2755.1}.

\bibitem[Liu et~al.(2025)Liu, Liu, Liang, Li, Liu, Wang, Wan, Zhang, and
  Ouyang]{liu2025flowgrpo}
Jie Liu, Gongye Liu, Jiajun Liang, Yangguang Li, Jiaheng Liu, Xintao Wang,
  Pengfei Wan, Di~Zhang, and Wanli Ouyang.
\newblock Flow-{GRPO}: Training flow matching models via online {RL}.
\newblock In \emph{Advances in Neural Information Processing Systems}, San
  Diego, CA, USA, 2025.
\newblock \doi{10.48550/arXiv.2505.05470}.

\bibitem[Jacq et~al.(2026)Jacq, Couairon, De~Bortoli, Berthet, Doucet, and
  Elie]{jacq2026rmmd}
Alexis Jacq, Guillaume Couairon, Valentin De~Bortoli, Quentin Berthet, Arnaud
  Doucet, and Romuald Elie.
\newblock Diffusion fine-tuning with rewarded moment matching distillation.
\newblock \emph{arXiv preprint arXiv:2606.30414}, 2026.
\newblock \doi{10.48550/arXiv.2606.30414}.

\end{thebibliography}

\end{document}